\documentclass[mlabstract]{jmlr}

\usepackage[hpos=300px,vpos=70px]{draftwatermark}
\usepackage{hyperref}
\usepackage{url}
\usepackage{booktabs}
\usepackage{multirow}
\usepackage{algorithm}
\usepackage{algorithmicx}
\usepackage{algpseudocode}
\usepackage{wrapfig}
\usepackage[normalem]{ulem}

\newcommand{\bc}[2]{\textcolor{#1}{\mathbf{#2}}}

\SetWatermarkText{\test}
\SetWatermarkScale{1}
\SetWatermarkAngle{0}

\usepackage{longtable}

\usepackage{booktabs}
\usepackage[load-configurations=version-1]{siunitx} 

\theorembodyfont{\upshape}
\theoremheaderfont{\scshape}
\theorempostheader{:}
\theoremsep{\newline}

\jmlrvolume{}
\firstpageno{1}

\jmlryear{2023}
\jmlrworkshop{Symmetry and Geometry in Neural Representations}

\title[Data Augmentations in Deep Weight Spaces]{Data Augmentations in Deep Weight Spaces}

  \author{\Name{Aviv Shamsian}$^{*1}$ \enspace
  \Name{David W. Zhang}$^{*3}$ \def\thefootnote{*}\footnotetext{Equal contribution}
  \Name{Aviv Navon}$^1$\enspace
  \Name{Yan Zhang}$^4$\enspace
  \Name{Miltiadis Kofinas}$^3$\\
  \Name{Idan Achituve}$^1$\enspace
  \Name{Riccardo Valperga}$^3$\enspace
  \Name{Gertjan J. Burghouts}$^5$\enspace
  \Name{Efstratios Gavves}$^3$\\
  \Name{Cees G. M. Snoek}$^3$\enspace
  \Name{Ethan Fetaya}$^1$\enspace
  \Name{Gal Chechik}$^{1,6}$\enspace
  \Name{Haggai Maron}$^{2,6}$\\
  \addr Bar Ilan University$^1$\enspace Technion$^2$\enspace University of Amsterdam$^3$\enspace Samsung\,-\,SAIT AI Lab, Montreal\,$^4$\enspace TNO\,$^5$\enspace NVIDIA$^6$
  }
  \vspace{-2mm}

\begin{document}

\maketitle
\begin{abstract}
Learning in weight spaces, where neural networks process the weights of other deep neural networks,  has emerged as a promising research direction with applications in various fields, from analyzing and editing neural fields and implicit neural representations, to network pruning and quantization. Recent works designed architectures for effective learning in that space, which takes into account its unique, permutation-equivariant, structure. 
Unfortunately, so far these architectures suffer from severe overfitting and were shown to benefit from large datasets. This poses a significant challenge because generating data for this learning setup is laborious and time-consuming since each data sample is a full set of network weights that has to be trained.
In this paper, we address this difficulty by investigating data augmentations for weight spaces, a set of techniques that enable generating new data examples \emph{on the fly} without having to train additional input weight space elements. We first review several recently proposed data augmentation schemes 
and divide them into categories. We then introduce a novel augmentation scheme based on the Mixup method. We evaluate the performance of these techniques on existing benchmarks as well as new benchmarks we generate, which can be valuable for future studies.
%
\end{abstract}

\section{Introduction}
Learning in deep weight spaces is the problem of training neural networks for processing the weights of other deep neural networks -- a problem that has recently gathered significant interest \citep{eilertsen2020classifying,unterthiner2020predicting,andreis2023set}.
The focus in recent work has largely been on the development of novel architectures that take into consideration the permutation symmetry of neurons \citep{navon2023equivariant,zhou2023permutation,zhou2023neural,zhang2023neural}. These architectures have demonstrated promising results on multiple benchmarks, significantly outperforming earlier na\"ive approaches. 

However, there is still a significant gap between the results obtained by such equivariant networks operating on Implicit Neural Representations (INRs) and those obtained by applying standard deep models such as CNNs and MLPs, which take as input the original image (or other raw signal) representation. For example, current state-of-the-art models for processing INRs achieve only 16\%  accuracy on the ModelNet40 3D shape classification benchmark, while neural networks operating on the same data, but represented using a point cloud, achieve over 90\% accuracy \citep{atzmon2018point,wang2019dynamic}. 

\begin{figure}[t]
    \centering

\begin{tabular}{cc}
  \includegraphics[width=0.35\textwidth]{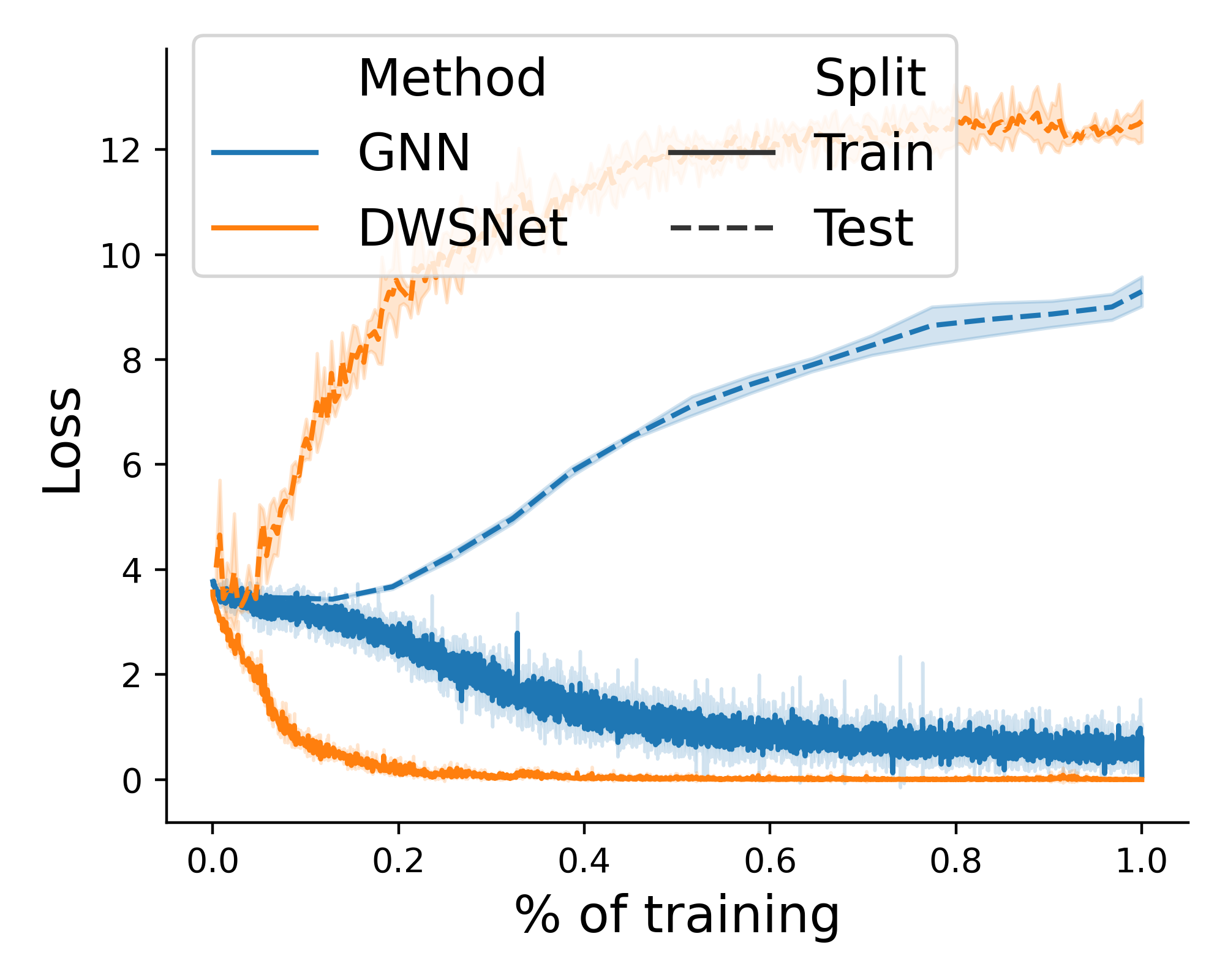}
  & 
  \includegraphics[width=0.35\textwidth]{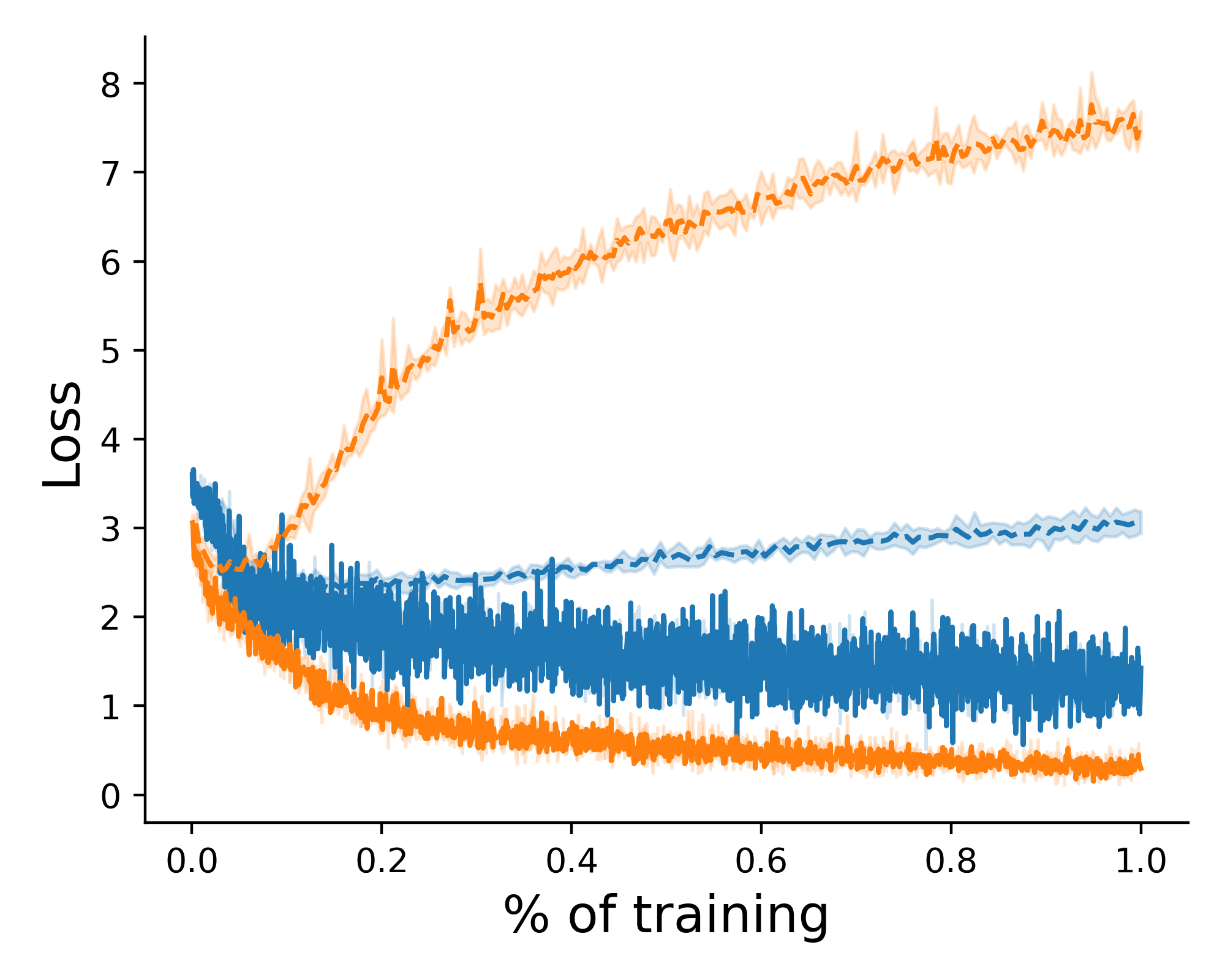}\\\vspace{-20pt}
  \label{fig:overfit_10views}
  \\
  (1) Training with 1 view per INR  &  (2) Training with 10 views per INR
\end{tabular}    
    \vspace{-2mm}
    \caption{\textit{Overfitting of weight space architectures on the ModelNet40 dataset}: We visualize train 
    and test 
    losses for DWSNets \citep{navon2023equivariant} and graph-based architectures \citep{zhang2023neural} on the ModelNet40 datasets with 1 
    or 10 
    trained input networks per point cloud (views). Notably, both methods tend to overfit early during training, even when using more data.}
    \label{fig:overfit}
\vspace{-28pt}
\end{figure}

To understand this problem, we first quantify the generalization error when learning with INRs: Figure~\ref{fig:overfit} demonstrates that existing architectures suffer from severe overfitting (see details in Section~\ref{sec:overfitting}).   
A possible explanation for this finding is that previous equivariant architectures account for the permutation symmetries in the weight space, but remain sensitive to other types of variability present in weight spaces. These include, for example, scaling transformations, weight perturbations, and more.
While it is possible to bridge the generalization gap by collecting more data, this is challenging when learning in weight spaces, where generating any data sample requires training a deep neural network.

To alleviate this problem and 
effectively increase the number of training examples without generating more data, we provide the first study of data augmentation schemes for weight spaces: we study simple transformations that can be applied to input samples  (weight space elements) to achieve more diversity while preserving the 
 functions represented by those weights. 
%
%
Data augmentation schemes are widely used and heavily studied for common data types like images \citep{shorten2019survey}. For weight spaces, augmentations are challenging and unexplored, in part, due to their symmetry structure. 
%

We first propose a taxonomy of known weight space augmentation schemes. (i) \emph{input-space augmentation}, transformations of weight space elements that reflect simple transformations in the input space, like rotating a 3D object by a linear transformation of its 3D INR; (ii) \emph{generic augmentations}, like adding noise, feature masking or dense feature Mixup (applying Mixup to the representation in the penultimate layer). See also \citep{zhang2017mixup,zhou2023neural}. In addition, we propose (iii) \emph{augmentations inspired by activation functions}, a novel family of augmentations that exploit activation symmetries.

We then develop \textit{Weight-Space Mixup}, a novel data augmentation scheme based on generalizing the Mixup method \citep{zhang2017mixup} to weight spaces. Unlike mixup for dense vectors and images, applying mixup directly to weight space elements is challenging. A crucial reason for that is that due to the permutation symmetries of weight spaces, the weights of two independently trained models are rarely aligned, and directly averaging them may not yield an appropriate model \citep{ainsworth2022git}. We address this difficulty and develop several variants of weight space mixup, building on recent works in weight-space alignment algorithms \citep{ainsworth2022git,pena2023re}.

Our results indicate that data augmentation schemes, and specifically our proposed Weight-Space Mixup method, can enhance the accuracy of weight space architectures by up to 18\%. This improvement is equivalent to reducing the required amount of generated data by almost 10$\times$, which saves countless hours of computation and electricity consumption. We also contribute two new benchmarks on both image and 3D modalities. 

\section{Overfitting in deep weight spaces}
\label{sec:overfitting}
In this section, we show that Weight space networks, which operate on INRs, significantly underperform compared to their counterparts that operate on the original data space (e.g., point clouds or images).
We attribute this performance gap primarily to an overfitting problem in deep weight spaces.
In order to validate this, we train two different weight space models DWSNet~\citep{navon2023equivariant} and GNN~\citep{zhang2023neural} on ModelNet40 INR datasets with 1 or 10 views per object, where by views we mean differently initialized INRs that fit the same object in the original dataset. 
In Figure~\ref{fig:overfit} (left panel), we observe that when training with a single view all runs start overfitting in the early stages of the training process at around $5-10\%$ of the total update steps. Training with 10 views per object (Figure~\ref{fig:overfit} right panel) somewhat alleviates the overfitting problem and can be seen as a type of data augmentation. We note that although the gap between the train and test error becomes smaller, the overfitting problem still remains. Furthermore, generating 10 views per object requires substantial computing time. Next, we consider alternative augmentation methods that can be applied directly to weight space elements.


\section{A taxonomy of augmentations for weight-space elements}
We present three families of augmentation schemes for weight spaces, and use these schemes to categorize previously proposed augmentation schemes. 


\paragraph{Input-space augmentations.}
Data augmentations like random rotations, translations, and scalings are frequently used when learning image and 3D data. As shown in \cite{navon2023equivariant}, in many cases, these augmentations can be applied to  INRs by applying the relevant geometric transformations to the input coordinates of the INR. As an example, rotating the object represented by an INR by a random rotation $R$ can be accomplished by replacing $W_1$, the first weight matrix of the INR, with $W_1R$. 

\paragraph{Generic augmentations.}
 General data augmentation techniques are augmentation techniques that can be applied to any type of data. This category includes several methods such as dropout~\citep{srivastava2014dropout}, which randomly deactivates a fraction of weights during training, quantile dropout, which removes weights with magnitudes below a defined threshold, and the addition of random Gaussian noise to the input weights.

\paragraph{Augmentations inspired by activation functions.}
\label{sec: weight_space_aug}
In many cases, activation functions induce symmetries that are not easy to incorporate into the weight space architecture. 
We propose three activation space augmentations that exploit this symmetry. For ReLU activation, we can arbitrarily scale the weights $\frac{1}{c}W_{i+1}\text{ReLU}(cW_ix{+}cb_i){+}b_{i{+}1} = W_{i+1}\text{ReLU}(W_ix{+}b_i){+}b_{i+1}$ with some $c{\in}\mathbb{R}^+$. In SIREN \citep{Sitzmann2020ImplicitNR}, the sinusoidal activation function induces two additional symmetries. First, since the function is odd we can negate the weight and biases of layer $i$ and the weight of the following layer $i{+}1$ as $W_{i+1}\text{Sine}(W_ix{+}b) = -W_{i+1}\text{Sine}(-W_ix{-}b)$. The second symmetry results from the shift of the phase in an even or odd multiple of $\pi$, more formally: $W_{i+1}\text{Sine}(W_ix+b) = (-1)^kW_{i+1}\text{Sine}(W_ix+b+k\pi)$. for $k\in\mathbb{Z}$. We incorporate these symmetries through random data augmentations and refer to them as \textit{SIREN negation} and \textit{SIREN bias} respectively.


\vspace{-2mm}
\section{Mixup in weight space}
Mixup \citep{zhang2017mixup} is a popular data augmentation technique where the basic idea is to randomly interpolate a pair of input images $x_1, x_2$ and their ground truth labels $y_1, y_2$ to create a new training example $(\lambda x_1 {+} (1-\lambda)x_2, \;\; \lambda y_1 {+} (1{-}\lambda)y_2)$. In the last few years, Mixup was successfully generalized to several data types such as point clouds and graphs \citep{chen2020pointmixup,achituve2021self,han2022g}.  
\vspace{-2mm}
\paragraph{Alignment and interpolation in weight spaces.} To design a mixup method for wight spaces, we first need to understand the weight space alignment problem: given two weight space elements $x_1=[W_1^{(l)}, b_1^{(l)}]$ and $x_2=[W_2^{(l)}, b_2^{(l)}]$, $l=1,\dots,M$, this problem seeks a sequence of permutations $p=(P_1,\dots, P_{M-1})$ that minimizes $\|x_1-p\cdot x_2\|$, where $p\cdot x_2$ applies the permutations to the weight vectors without changing the underlying function, as in Equation 5 in \cite{navon2023equivariant}. Intuitively, this problem seeks permutations such that the weights of these networks are as close as possible when compared directly. Several recent works \citep{entezari2022the,ainsworth2022git,pena2023re,navon2023equivariant} have shown that the interpolation between a weight vector $x$, to the optimally permuted version of the other vector $p \cdot x'$ has a property called \emph{linear mode connectivity}, which states that the loss value on this path is only marginally worse compared to its endpoints. This is in contrast to weights obtained from the direct interpolation between $x,x'$ which produces a significant increase in this loss.

\vspace{-2mm}
\paragraph{Weight-space mixup.} 
The naive (standard) \emph{weight-space mixup} is formally defined as an interpolation between two weight space samples $[W_1^{(l)}, b_1^{(l)}]$ and $[W_2^{(l)}, b_2^{(l)}]$ with $\lambda\sim\mathcal{U}(0,1)$:
$ W^{(l)} = \lambda W_1^{(l)} {+} (1{-}\lambda) W_2^{(l)}, \;\; b^{(l)} = \lambda b_1^{(l)} {+} (1{-}\lambda) b_2^{(l)}$, where the weight parameter $\lambda$ is randomly drawn from a uniform distribution.

Next, we define the \emph{randomized weight space mixup} in which random permutations are applied to one of the input weights before mixing two samples.
While the weights, in this case, are still not aligned (with high probability), we do get a much greater degree of diversity than we would obtain with the standard approach.

Lastly, we define \emph{matching based weight space mixup} where we use a sequence of permutation matrices $p$ to first align the weights and then perform the interpolation. As the weight space alignment problem is NP-hard,  we obtain an approximate alignment using the \emph{Weight Matching} algorithm suggested by \citep{ainsworth2022git}.

\vspace{-2mm}
\section{Experiments}
\label{sec:experiments}
We evaluate various weight-space augmentations for classifying INRs. Specifically, we create INR datasets for ModelNet40 (3D point clouds) and FMNIST (2D greyscale images). We generate 10 different INRs -- referred to as 10 different views -- for each example in the original dataset. We compare each augmentation individually for 1 and 10 views and use DWS \citep{navon2023equivariant} and GNN \citep{zhang2023neural} as weight-space architectures. We report the average accuracy and standard deviations for $3$ random seeds.
More details on the experimental setup and data generation processes are in Appendix~\ref{app:exp_details},~\ref{app:datasets}.



\noindent Table~\ref{tab:inrs} showcases the effectiveness of on-the-fly weight space data augmentation schemes. Notably, \textbf{Mixup augmentations with a single view are comparable to training with 10$\times$ more data} on both datasets: ModelNet40 and FMNIST. Furthermore, data augmentation is still effective with 10 views.
Augmentations applied in the input space, which are limited to the first layer of the INR, are less effective compared with other types of augmentations that modify the weights in all the layers.
Overall, the effectiveness of input-space and generic augmentations varies between the models and also between the datasets.
In contrast, Weight Space Mixup provides consistent improvements, with the alignment-based version frequently outperforming other variants. 
\vspace{-2mm}

\begin{table}[t]
\scriptsize
\centering
\caption{\textit{ModelNet40 and FMNIST results}: test accuracy results for $1$ and $10$ views.} 
\vspace{5pt}
\begin{tabular}{lcccccc}
\toprule
Augmentation type & Model & \multicolumn{2}{c}{ModelNet40} &  & \multicolumn{2}{c}{FMNIST} \\ \cmidrule(lr){3-4} \cmidrule(lr){6-7} 
 &  & 1 View & 10 View &  & 1 View & 10 View \\

\midrule 
No augmentation  & DWS  &  $16.17 \pm 0.25$ & $30.25 \pm 0.95$ && $68.30 \pm 0.62$ & $76.01 \pm 1.20$  \\
No augmentation  & GNN  &  $8.82 \pm 1.08$  & $34.51 \pm 1.24$ && $68.84 \pm 0.41$ & $79.58 \pm 3.01$ \\
\midrule
Translate        &  DWS  &   $18.18 \pm 0.97$  & $31.17 \pm 0.02$ && $67.90 \pm 0.24$ & $77.61 \pm 0.36$ \\
Rotation         &  DWS  &   ---   & --- && $68.55 \pm 0.28$ & $77.04 \pm 0.47$ \\
Scale            &  DWS  &   $16.41 \pm 0.57$   & $30.54 \pm 0.72$ && $67.99 \pm 0.14$ & $75.77 \pm 1.09$ \\
Gaussian noise   &  DWS  &   $14.10 \pm 0.71$   & $25.31 \pm 1.78$ && $68.53 \pm 0.09$ & $77.60 \pm 0.13$ \\
SIREN bias       &  DWS  &   $4.69 \pm 0.10$   & $4.90 \pm 0.01$ && $58.20 \pm 0.01$ & $62.21 \pm 0.55$ \\
SIREN negation   &  DWS  &   $20.14 \pm 0.98$   & $32.31 \pm 0.70$ && $71.40 \pm 0.29$ & $77.71 \pm 1.38$ \\
Dropout          &  DWS  &   $11.43 \pm 2.44$   & $14.71 \pm 1.14$ && $68.48 \pm 0.14$ & $75.57 \pm 1.91$ \\
Quantile dropout &  DWS  &   $15.13 \pm 2.45$   & $29.88 \pm 0.62$ && $68.72 \pm 0.27$ & $76.22 \pm 0.72$ \\
Translate        &  GNN  &   $8.17 \pm 0.81$   & $34.93 \pm 1.31$ && $70.17 \pm 1.26$ & $\bc{black}{83.83 \pm 0.25}$ \\
Rotation         &  GNN  &   ---   & --- && $69.35 \pm 2.18$ & $83.72 \pm 1.14$ \\
Scale            &  GNN  &   $8.58 \pm 0.65$   & $34.70 \pm 5.19$ && $68.96 \pm 1.46$ & $83.67 \pm 0.19$ \\
Gaussian noise   &  GNN  &   $9.06 \pm 0.27$   & $32.82 \pm 1.14$ && $77.55 \pm 0.33$ & $81.28 \pm 0.50$ \\
SIREN bias       &  GNN  &   $11.63 \pm 2.48$   & $34.32 \pm 1.57$ && $68.09 \pm 0.49$ & $77.20 \pm 1.03$ \\
SIREN negation   &  GNN  &   $11.41 \pm 3.22$   & $37.93 \pm 2.26$ && $72.74 \pm 4.29$ & $82.36 \pm 3.66$ \\
Dropout          &  GNN  &   $8.10 \pm 0.43$   & $18.04 \pm 1.24$ && $68.55 \pm 1.21$ & $79.72 \pm 1.35$ \\
Quantile dropout &  GNN  &   $8.12 \pm 0.85$   & $34.36 \pm 1.14$ && $69.96 \pm 2.08$ & $83.78 \pm 0.76$ \\
\midrule 
MixUp            &  DWS  &   $26.96 \pm 0.91$   & $31.92 \pm 0.37$ && $74.36 \pm 1.17$ & $78.58 \pm 0.20$ \\
MixUp + random perm. &  DWS &  $26.62 \pm 0.18$   & $\bc{black}{33.55 \pm 1.40}$ && $73.89 \pm 0.89$ & $78.04 \pm 1.02$ \\
Alignment + MixUp    &  DWS &  $\bc{black}{27.40 \pm 0.97}$   & $33.33  \pm 0.43$ && $\bc{black}{75.67 \pm 0.36}$ & $\bc{black}{79.41 \pm 0.56}$ \\
MixUp            &  GNN  &   $20.45 \pm 3.82$   & $42.25 \pm 3.83$ && $\bc{black}{80.18 \pm 0.59}$ & $82.20 \pm 0.52$ \\
MixUp + random perm. &  GNN &  $24.46 \pm 2.92$   & $41.67 \pm 4.55$ && $78.45 \pm 2.29$ & $82.24 \pm 0.68$ \\
Alignment + MixUp    &  GNN &  $\bc{black}{26.88 \pm 1.75}$   & $\bc{black}{42.83 \pm 4.18}$ && $78.80 \pm 2.12$ & $82.94 \pm 0.31$ \\
\bottomrule

\end{tabular}
\label{tab:inrs}
\vspace{-4mm}
\end{table}
\vspace{-1mm}
\section{Conclusion}
\vspace{-1mm}
This paper examines the overfitting issue associated with weight space architectures and proposes novel weight space augmentation techniques that mitigate this issue and enhance model performance. Notably, our experiments demonstrate that training with these augmentations has comparable results to training with substantially larger datasets.

\noindent \textbf{Limitations.} It is important to note that weight space augmentations may vary in effectiveness across different datasets and tasks, which requires further investigation. 
In addition, some augmentations, such as Mixup with alignment, may require additional computational overhead that may be prohibitive in some resource-constrained environments.

\noindent \textbf{Acknowledgements.} HM is the Robert J. Shillman Fellow, and is supported by the Israel Science Foundation through a personal grant (ISF 264/23) and an equipment grant (ISF 532/23).

\newpage
\bibliography{pmlr-sample}

\newpage
\appendix

\section{Previous work}

\paragraph{Learning in deep weight spaces}
Recently there has been a growing interest in applying deep learning architectures directly to neural network weights. The attention to this domain was brought up by studies that presented weight-space related datasets~\citep{Dupont2022FromDT, Schrholt2022ModelZA}. Others~\citep{Sitzmann2020ImplicitNR} showed the possibility of representing a data point as an implicit neural representation on several tasks from classification to generation. These datasets raised the motivation to study how to learn in deep weight spaces. Early methods proposed using simple architectures such as MLPs and transformers to predict test errors or the hyperparameters that were used for training input networks \citep{eilertsen2020classifying, unterthiner2020predicting}. 
Recently, \cite{navon2023equivariant} presented the first neural architecture that accounts for natural permutation symmetries of weight spaces and demonstrated significant performance improvements over prior methods. \cite{zhou2023permutation} proposed a similar approach, which was later enhanced by the addition of attention mechanisms \citep{zhou2023neural}.  Finally, \cite{zhang2023neural} proposed a GNN architecture to process neural networks modeled as computational graphs.

\paragraph{Data augmentation in deep learning}
Data augmentation is an essential technique in deep learning that plays a crucial role in mitigating overfitting while enhancing the generalization capabilities of NNs. Data augmentation helps the model learn more robust and invariant features. Various techniques are employed to achieve this, such as geometric transformations like rotation, scaling, and translation which make the model robust to changes in object orientation and position. Additionally, color-based augmentations like brightness adjustments, contrast changes, and color jittering contribute to improved generalization by increasing the model's tolerance to variations in lighting conditions. Dropout~\citep{srivastava2014dropout} is another prominent data augmentation approach that randomly deactivates weights during training. These methods collectively enhance the model's ability to generalize from limited training data and reduce the risk of overfitting, resulting in more robust and accurate deep learning models. 

\paragraph{Mixup}
Mixup is a data augmentation method that blends two or more training samples to create new synthetic instances~\citep{cao2022survey}. Mixup~\citep{zhang2017mixup} operates by taking a weighted linear combination of two input samples, where both the input data and their corresponding labels are mixed. The resulting mixed data point contains characteristics of both original samples, effectively generating a smooth interpolation between them. Various mixup variants have been proposed, including CutMix~\citep{yun2019cutmix}, which combines two images by cutting and pasting rectangular regions; and AugMix~\citep{hendrycks2019augmix}, which applies multiple augmentation operations before mixing to further diversify the dataset. Recently, several works~\citep{ling2023graph} proposed performing Mixup after first aligning the feature or input space, resulting in smoother interpolation between the mixed objects.

\section{Datasets}
\label{app:datasets}
The increasing usage of INRs in many machine-learning domains, specifically in images and 3D objects, raises the need for INR benchmarks. Implicit representations, such as neural radiance fields and neural implicit surfaces, offer a more flexible and expressive way to model complex 3D scenes and objects. However, as these techniques gain traction, it becomes crucial to establish standardized benchmarks to assess and compare the performance of architectures designed for weight space data. To address this issue, we present new INR classification benchmarks based on ModelNet40~\citep{wu20153d} and Fashion-MNIST~\citep{Xiao2017FashionMNISTAN} datasets. We use the SIREN~\citep{Sitzmann2020ImplicitNR} architecture, i.e. MLP with sine activation, and fit each data point in the original dataset. To negate the possibility of canonical representation that may lead to globally aligned data representation, we randomly initialize the weights for every generated INR. 
In the case of ModelNet40, INRs are generated through training an MLP to accurately predict the signed distance function values of a 3D object given a set of 3D point clouds. For Fashion-MNIST an MLP is trained to map from the 2D xy-grid to the corresponding gray level value in the original image. 
We fit $10$ unique INRs, namely views, per sample in the original dataset resulting in a total of $123K$ and $700K$ INRs for ModelNet40 and Fashion-MNIST respectively.

\section{Experimental Details}
\label{app:exp_details}
\paragraph{DWS.}
In all experiments, we use DWS~\citep{navon2023equivariant} network with $4$ hidden layers and hidden dimension of $128$. We optimized the network using a $5e-3$ learning rate with AdamW~\citep{Loshchilov2017FixingWD} optimizer.

\paragraph{GNN.}
For the GNN, we use the version of Relation Transformer presented in \citet{zhang2023neural} with 4 hidden layers, node dimension of $64$, and edge dimension of $32$. We optimized the network using a $1e-3$ learning rate with AdamW~\citep{Loshchilov2017FixingWD} optimizer and a $1000$ steps warmup schedule.

\paragraph{General.}
We optimized the model for $250$ epochs for the ModelNet40 experiments and $300$/$100$ epochs for the FMNIST 1/10 views respectively. Additionally, we utilize the validation set for early stopping, i.e. selecting the best model w.r.t validation accuracy. We repeat all experiments using $3$ random seeds and report the average classification accuracy along with the standard deviation.

\section{Weight space augmentation details}
\paragraph{input space-based augmentations.}
Similar to rotation, scaling the coordinates by a factor $s$ is equivalent to scaling the weights $W_1(sx)=(W_1s)x$. Furthermore, a translation by an offset $t$ can be absorbed into the bias $W_1(x+t)+b_1=W_1x+(W_1t+b_1)$. These augmentations are natural, but they only change the parameters of the first layer, so their effectiveness may be limited.
\paragraph{General data augmentations.}
Dropout augmentation sets a parameter to $0$ with probability $p_{\text{drop}}$. Quantile-based dropout first computes a threshold based on which it zeroes out the $q$-th quantile that is closest to $0$. Gaussian noise augmentation adds Gaussian noise to all parameters with the standard deviation set in relation to the layer's standard deviation between the parameters. 

\begin{wrapfigure}{r}{0.5\textwidth}
    \includegraphics[width=0.48\textwidth, height=.65\textheight]{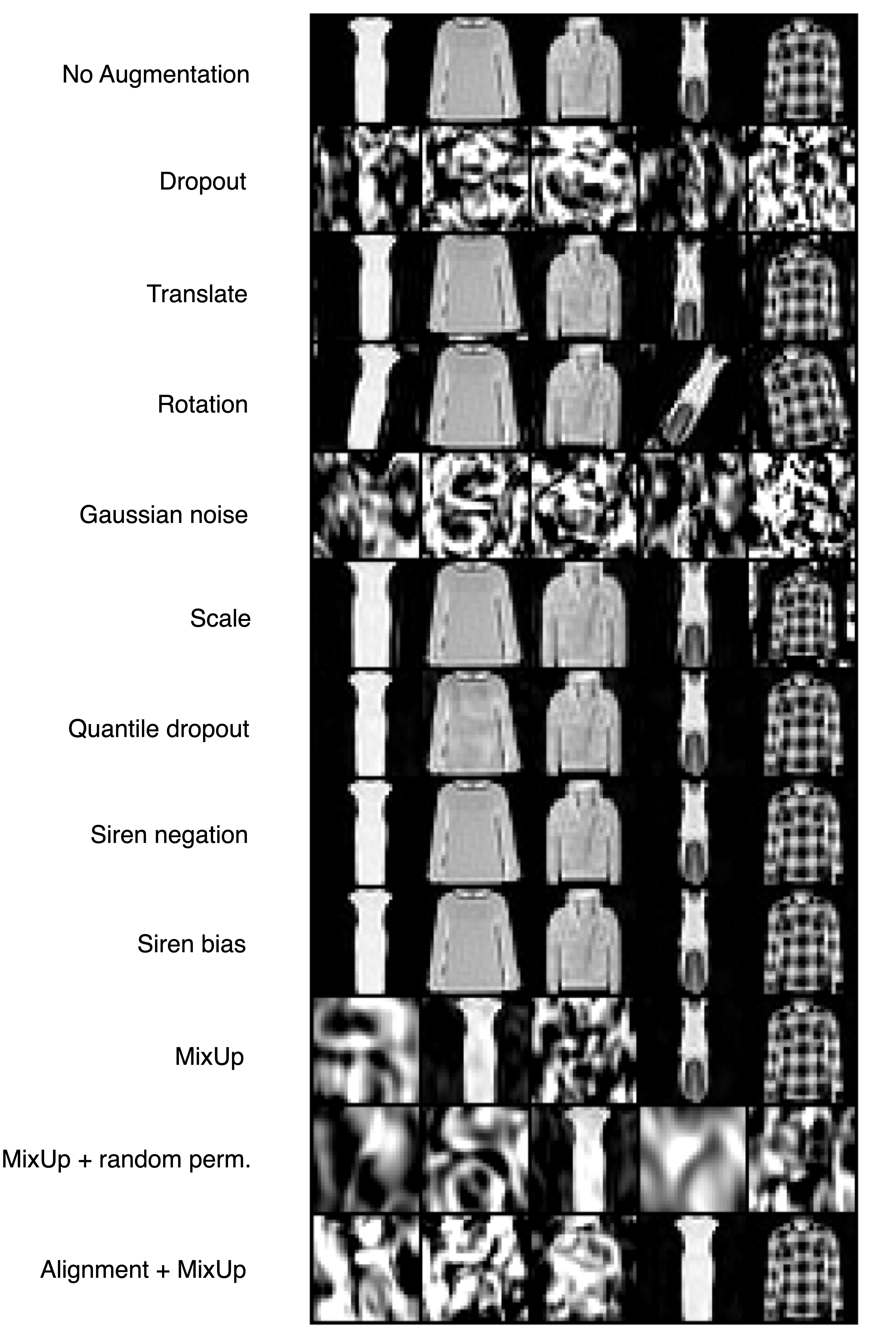}
    \caption{\emph{Visualizing augmentations}: Illustrating the influence of weight space augmentations on the image Space.}
    \label{fig:augmentation_vis}
\end{wrapfigure}
\newpage
\section{Datasets generation}

\emph{Fashion-MNIST INRs.} We fit an INR to each image in the original dataset. We split the INRs dataset into train, validation, and test sets of sizes 55K, 5K, and 10K respectively. Each INR is a $3$-layer MLP network with a $32$ hidden dimension, i.e., $3 \xrightarrow{} 32 \xrightarrow{} 32 \xrightarrow{} 1$. We train the INRs using the Adam optimizer for $1K$ steps with a learning rate of $5e-4$. When the PSNR of the reconstructed image from the learned INR is greater than $40$, we use early stopping to reduce the generation time.

\emph{ModelNet40.}
We use the original split presented in~\cite{wu20153d} and fit an INR for each data sample. We start by converting the mesh object to a signed distance function (SDF) by sampling $250K$ points near the surface. Next, we fit a 5-layer INR with a hidden dim of $32$, i.e., $3 \xrightarrow{} 32 \xrightarrow{} 32 \xrightarrow{} 32 \xrightarrow{} 32 \xrightarrow{} 1$ by solving a regression problem. Given a 3 dimensional input, the INR network predicts its SDF. For the optimization, we use AdamW optimizer with $1e-4$ learning rate and perform $1000$ update steps. 

\section{Visualization of reconstructed images}
Here, we investigate how augmentations applied to INR's weights affect the image reconstructed from the augmented INR. We consider the FMNIST INR dataset and apply all the augmentations we explored in this paper. Then we plot the reconstructed images, shown in Figure~\ref{fig:augmentation_vis}.

\end{document}